\pdfoutput=1

\documentclass[11pt]{article}

\usepackage[final]{acl}

\usepackage{times}
\usepackage{latexsym}
\usepackage{CJKutf8}

\usepackage[T1]{fontenc}

\usepackage[utf8]{inputenc}

\usepackage{microtype}

\usepackage{inconsolata}

\usepackage{amsmath}

\title{LLM-based Code-Switched Text Generation for \\ Grammatical Error Correction}

\author{Tom Potter\thanks{\hspace{4pt}Work completed whilst at King's College London.} \\
  University of Manchester\\
  \texttt{thomas.potter@postgrad.manchester.ac.uk}\\\And
  Zheng Yuan \\
  King's College London \\
  \texttt{zheng.yuan@kcl.ac.uk} \\}

\begin{document}
\maketitle
\begin{abstract}
With the rise of globalisation, code-switching (CSW) has become a ubiquitous part of multilingual conversation, posing new challenges for natural language processing (NLP), especially in Grammatical Error Correction (GEC). This work explores the complexities of applying GEC systems to CSW texts. Our objectives include evaluating the performance of state-of-the-art GEC systems on an authentic CSW dataset from English as a Second Language (ESL) learners, exploring synthetic data generation as a solution to data scarcity, and developing a model capable of correcting grammatical errors in monolingual and CSW texts. We generated synthetic CSW GEC data, resulting in one of the first substantial datasets for this task, and showed that a model trained on this data is capable of significant improvements over existing systems. This work targets ESL learners, aiming to provide educational technologies that aid in the development of their English grammatical correctness without constraining their natural multilingualism.
\end{abstract}

\section{Introduction}
Code-switching (CSW), the practice of fluidly alternating between two or more languages in conversation, has become commonplace in recent years. This linguistic phenomenon, emerging as a natural consequence of multilingualism, is now widely accepted in social and professional settings \cite{YOW2018}. Many works have highlighted the utility and cultural importance of CSW in general conversation \cite{BEATTY2020, FALBO2021}. Further research indicates that these advantages extend to language learning, with CSW offering many pedagogical benefits. These include increasing students' access to content and improving their confidence. \citet{NGUYEN2022} discuss the mechanisms for this, where students use a familiar language to grasp foreign, complex concepts. CSW can also serve as a scaffolding tool, helping to bridge gaps in a student's comprehension of a language and enabling them to build upon existing knowledge. These benefits reduce the barriers between a student and their target language and help promote a learning environment conducive with active exploration and deeper understanding. Therefore, it is essential that English as a Second Language (ESL) learners are not penalised for expressing their cultural identity through CSW. Grammatical error correction (GEC) is the task of automatically detecting and correcting errors in text. Research on GEC for CSW text remained largely unexplored. \citet{chan-etal-2024-grammatical} were the first to demonstrate that exposing a sequence-tagging GEC model to CSW text during the training process improves performance compared to a monolingual system. However, further work is essential to ensure language technology is inclusive and reflective of real-world linguistic practices.
Figure \ref{fig:csw_gec_example} shows two examples of CSW from our target population with their grammatical corrections.\footnote{The definition of CSW is a subject of ongoing debate. Throughout this work, we use the term CSW to refer specifically to the type of language mixing exhibited by ESL learners.}




\begin{figure}[t]
\small
    \centering
        \begin{minipage}{0.9\linewidth}
            \emph{Example 1:}\begin{CJK}{UTF8}{min}According to the test, [\textcolor{red}{lacks in me} $\rightarrow$ \textcolor{green}{my shortcomings}] are 靴下 and ご主人様.\end{CJK}
            
        \emph{Example 2}: \begin{CJK}{UTF8}{min}When we [\textcolor{red}{call} $\rightarrow$ \textcolor{green}{say}] ダッシュボード, do we actually mean a glove compartment in English?\end{CJK}
        \end{minipage}
    \caption{Examples of GEC in ESL learner language.}
    \label{fig:csw_gec_example}
\end{figure}



Despite significant advancements in GEC in recent years, a gap persists in addressing CSW texts, with monolingual GEC datasets labelling CSW as a type of error \cite{NGUYEN2022}. There are several reasons for this, the most prominent being the scarcity of high-quality training data, a problem that plagues monolingual GEC systems. The unique linguistic features of CSW, including its variable syntax, semantics and pragmatics, add additional complexity to this task. Monolingual seq2seq GEC models, e.g. T5 \cite{ROTHE2021}, struggle with CSW text as they fail to represent the non-English inputs, resulting in their inability to output the CSW text. On the other hand, multilingual seq2seq models and edit-based GEC models like GECToR \cite{OMELIANCHUK2020} can handle CSW text but struggle with the ambiguity present at language switching points. This ambiguity challenges the models’ ability to accurately correct the text.

This paper aims to bridge this gap. Firstly, to address the data scarcity issue, we propose a method for generating high-quality synthetic CSW GEC data, using which we produce, to our knowledge, one of the first substantial datasets labelled for this task\footnote{This dataset is available on \href{https://github.com/tpotterer/Synthetic-CSW-Text-for-GEC}{GitHub.}}. Secondly, we train a token classification-style GEC system, tailored to correct errors in texts produced by ESL learners. This demographic is significant for our study as they not only present consistent CSW patterns but also stand to benefit greatly from a GEC system capable of handling CSW text.

\section{Data}

\subsection{Genuine CSW GEC Dataset}
One of the only datasets labelled for GEC which does not remove CSW text, is the Lang-8 dataset \cite{lang8_dataset}, sourced from the Lang-8 language learning platform. This dataset, when filtered to contain entries where CSW is present, offers a foundation of authentic data, comprises 5,875 pairs of ungrammatical and corrected sentences across 6 CSW language pairs: English-Japanese (81.9\%), English-Korean (13.0\%), English-Traditional Chinese (3.4\%), English-Russian (1.2\%), English-Thai (0.5\%) and English-Arabic (0.1\%). 

The crowd-sourced nature of Lang-8 required manual validation to ensure accuracy. We tasked an annotator with the responsibility of verifying the original corrections in the dataset, as well as combing for missed errors, incorrect annotations and over-annotations.

\subsection{Synthetic CSW GEC Data Generation}
Given the small size of the available CSW GEC dataset, we introduced a 2-step approach to synthetic CSW GEC data generation. First, we generated grammatically correct CSW sentences. This is followed by the introduction of errors.

\begin{table*}[!h]
\footnotesize
\begin{center}
\begin{tabular}{l|rrrr}
\textbf{Metric} & \textbf{Genuine CSW} & \textbf{LLM CSW} &\textbf{Translation CSW}& \textbf{Corpus CSW}\\ [0.5ex]
\hline
CMI & 15.52 & 16.14 & 27.81 &11.42 \\
M-Index & 0.007 & 0.004& 0.015&0.006\\
I-Index & 0.21 & 0.21 & 0.30 &0.20\\
Burstiness & -0.07 & -0.04&0.03&-0.11\\
CF1 & 6.38 & 5.82 & 17.13&2.55\\
CF2 & 19.77 & 19.03&31.11&16.04\\
CF3 & 18.34 & 17.61&30.05&14.20\\
\end{tabular}
\end{center}
\caption{Quantitative Description of the Genuine and Generated CSW Datasets Using Various CSW Metrics.}
\label{table:csw-genuine-metrics}
\end{table*}

\subsubsection{Step 1: CSW Text Generation}

Three different synthetic data generation techniques have been explored to generate CSW data.

\paragraph{Translation-based CSW Text Generation} required a monolingual corpus, a machine translation (MT) algorithm, and a sentence parser. To generate a CSW utterance, we used the Stanford Parser v4.5.4\footnote{This can be downloaded from the \href{https://stanfordnlp.github.io/CoreNLP/history.html}{CoreNLP website}.} \cite{manning-etal-2014-stanford} to build a syntactic parse tree. We then randomly selected and translated a subtree using the ArgosTranslate MT package\footnote{We used ArgosTranslate v1.8.0, available on \href{https://github.com/argosopentech/argos-translate}{GitHub}.} \cite{ArgosTranslate, klein-etal-2017-opennmt}. This method generates plausible CSW text. However, performance is dependent on the strength of the parsing and translation algorithms; and the style of language within the corpus. To approximate the style of our authentic CSW text, we used corrected monolingual sentences from the Lang-8 corpus.

\paragraph{Parallel Corpus-based CSW Text Generation} avoids the need for a translation algorithm. Instead, we used the same Stanford Parser, this time with Spanish, French and German configurations; and the AWESOME word-alignment model\footnote{The authors shared their model on \href{https://huggingface.co/aneuraz/awesome-align-with-co}{HuggingFace}.} \cite{dou-neubig-2021-word}, to identify parts of parallel corpora labelled for MT with similar syntactic structure. For this method, we use the Europarl corpus \cite{epppc} due to the grammatical quality of its English component. Under the Equivalence Constraint Theory \cite{RIZVI2021}, these areas are where CSW is likely to take place. We, therefore, randomly chose overlapping subtrees as candidates for injection of non-English text. Although this method does not require MT, it is reliant on performant word-alignment and parsing systems; these are rare for many languages.

\paragraph{LLM Prompting-based CSW Text Generation}
The other methods of generating CSW text rely on injecting a second language into existing monolingual corpora. Hence, they are not able to recreate one of the main switching styles shown by ESL learners - CSW as a genuine pragmatic strategy. A common reason for this style of switching is when quoting another language. It is difficult to recreate this style using a monolingual foundation as sentences like \textit{The Japanese word for ``dog'' is ``\begin{CJK}{UTF8}{min}犬\end{CJK}''} seldom appear in authentic monolingual corpora.

To generate diverse CSW texts without relying on existing corpora or inaccurate alignment algorithms, we leveraged the strong general knowledge of Large Language Models (LLMs). We demonstrated that OpenAI’s GPT-3.5 \cite{gpt35} can create high-quality CSW sentences when shown examples of authentic utterances. Along with genuine CSW texts, we supplied a one-shot example of how to use the switching styles of an existing CSW text to generate a new sentence.\footnote{The full prompt can be seen in Appendix \ref{sec:llm_prompt}.}

\paragraph{Comparison of Synthetic CSW Text} We used several CSW metrics to quantify the qualities of CSW texts: Code Mixing Index (CMI) \cite{DAS2016}, Multilingual Index (M-Index) \cite{BARNETT2000}, Probability of Switching (I-Index) \cite{GUZMAN2017}, Burstiness \cite{GOH2008}, and Complexity Factor (CF1-3) \cite{GHOSH2017}. Table \ref{table:csw-genuine-metrics} shows the value of each metric for our genuine CSW dataset, as well as for these 3 synthetic CSW datasets. We can see that the LLM prompting-based dataset was superior in its similarity to the authentic CSW data. Using this method, we generated a corpus of 73,293 utterances covering over 20 English language pairs, including the 6 language pairs in the original dataset.\footnote{The LLM does not always generate the language pairs we ask for. However, these sentences are still included in the dataset categorised under their actual language pair.}

\subsubsection{Step 2: Synthetic Error Generation}


Several works have shown the effectiveness of rule-based error injection for GEC data generation. Many use the PIE-synthetic dataset \cite{AWASTHI2019}, a perturbed version of the 1BW corpus \cite{1bw_dataset}. For each sentence, the authors introduce between 0 and 4 errors of random type. We extended this work by introducing a new subset of error types that are not only more common in ESL learners, but also are areas where the SOTA performance collapses when faced with CSW text: noun, pronoun, word order, determiner, and punctuation errors.\footnote{Error type analysis is presented in Appendix \ref{sec:errant_analysis}.}

To increase the diversity of errors, we adopted a second style of error injection, Backtranslation \cite{STAHLBERG2021}. By swapping the source and target sentences of a monolingual dataset, we trained a GECToR-based system to induce errors in our synthetic CSW sentences.

Using these methods, we created two datasets: Syn-CSW PIE and Syn-CSW Rev-GECToR. After removing pairs containing no injected errors, we are left with 70,180 and 18,159 sentences each. 

\begin{table*}[!h]
\footnotesize
\begin{center}
\begin{tabular}{l |c c c c c c}
\textbf{Model} & \multicolumn{3}{c}{\textbf{BEA-2019 Test}} & \multicolumn{3}{c}{\textbf{Genuine CSW}} \\
& \textbf{P} & \textbf{R} & $\text{\textbf{F}}_{0.5}$ & \textbf{P} & \textbf{R} & $\text{\textbf{F}}_{0.5}$ \\
\hline
\textbf{Existing GEC systems} & \multicolumn{6}{l}{} \\
GECToR & 77.88 & 53.07 & 71.22 & 71.14 & 27.08 & 53.67 \\
T5-Small & 62.03 & 47.19 & 58.34 & 11.70 & 24.98 & 13.09 \\
\hline
\textbf{Our CSW GEC systems} & \multicolumn{6}{l}{} \\
Stage 1 & 67.23 & 53.88 & 64.05 & 66.15 & 26.04 & 50.57 \\
Stage 2 & 72.64 & 51.73 & 67.20 & 65.41 & 29.93 & 52.87 \\
Stage 3 & 74.32 & 53.40 & 68.92 & 84.66 & 22.92 & 55.02 \\
Inference Tweaks & 69.01 & 58.40 & 66.59 & 76.02 & 38.67 & 63.71 \\
\end{tabular}
\end{center}
\caption{ERRANT-based Precision, Recall and $F_{0.5}$ Scores of Baselines and Our Model Throughout Training}
\label{table:model_scores}
\end{table*}

\section{CSW GEC Systems}
For our GEC system targeting CSW texts, we chose a GECToR model \cite{OMELIANCHUK2020}, with a RoBERTa-base foundation, due to its proven efficacy with limited training data and stronger performance on CSW texts compared to seq2seq models. We added a new CSW class to the error detection head, adding the ability to detect CSW tokens.

Following \citet{TARNAVASKYI2022}, we used a 3-stage training schedule. In the first, we used the same distilled 1BW corpus, and added all our synthetic CSW GEC data. For the second, we used several GEC datasets: NUCLE \cite{nucle_dataset}, FCE \cite{fce_dataset}, W\&I Locness \cite{bea-2019_dataset}, Lang-8 and our 2 synthetic CSW datasets. As our genuine CSW dataset is a subset of the Lang-8 corpus, we checked and removed any duplicates. Following previous works, we finished training using the W\&I Locness dataset due to its superior quality. In this final stage, we added a sampled subset of our synthetic CSW sentences and 90\% of our authentic CSW data, ensuring exposure to synthetic and genuine CSW text. At each stage, we reserved 5\% of the data for validation.\footnote{An exact breakdown of contributions by each dataset is given in Appendix \ref{sec:train_data}.} Finally, we tuned inference parameters using a grid-search to optimise the $F_{0.5}$ on the final validation set. By beginning with pre-training on large amounts of lower-quality data in the early stages, this multi-stage learning process allows the model to first build a robust GEC foundation before refining it with high quality data in the latter stages. This approach allows the model to learn incrementally, reducing the risk of the model being overwhelmed by the complexity of the task from the outset.

\section{Results and Analysis}
\subsection{Baseline Comparisons}
We compared our model against two well-established systems: a RoBERTa-base GECToR model \cite{OMELIANCHUK2020}, with near SOTA performance on the BEA-2019 test set \cite{bea-2019_dataset}, and a seq2seq T5 model \cite{ROTHE2021}. To assess these models, we evaluated their performance on the BEA-2019 test set and the remaining 10\% of our authentic CSW data. The ERRANT \cite{BRYANT2017} GEC evaluation results, as outlined in Table \ref{table:model_scores}, demonstrate a clear degradation in performance when these two systems are applied to CSW texts. The ERRANT toolkit detects and classifies edits between source and target sentence pairs into predefined error categories. It enables the comparison of a proposed set of edits with a reference set, providing a way of calculating metrics, such as precision and recall, across these categories.

\subsection{Detailed Model Performance}

The progression of our model throughout training provided insights into its evolving capabilities and effectiveness of our synthetic data. We monitored several metrics, including the ERRANT precision, recall and $F_{0.5}$ score, for the BEA-2019 test set and the remaining unused 10\% of our genuine CSW dataset. These metrics, as displayed in Table \ref{table:model_scores}, indicate a steady improvement in the ability to handle CSW texts. Notably, the performance on the CSW dataset shows a significant leap in the final stages, where the contribution of our synthetic dataset is largest.  This improvement in CSW text handling did slightly compromise the model’s performance on monolingual GEC tasks, as seen on the BEA-2019 test set. This suggests a trade-off inherent in specialising the model for CSW contexts. However, our model remains competitive amongst SOTA monolingual GEC systems of its size.

Three illustrative examples of our model's corrections, taken from the CSW test set, can be seen in Figure \ref{fig:csw_examples}. The first example demonstrates a case where the model has correctly identified all of the changes required, including the incorrect capitalisation of a word, a missing word, and some missing punctuation. The second example shows a ``near miss''; here, the model has correctly identified the majority of the changes required but dropped the ``I'' whilst rearranging the start of the sentence. Finally, the third example presents a scenario where the model has fallen slightly short, failing to recognise the need for ``were'' instead of ``was'' in this hypothetical context.

\begin{figure*}[t]
\small
    \centering
        \begin{minipage}{0.9\linewidth}
        \emph{\textbf{Gold Correction 1:}}\begin{CJK}{UTF8}{mj} We have many [\textcolor{red}{New} $\rightarrow$ \textcolor{green}{new}] words for [\textcolor{red}{$\emptyset$} $\rightarrow$ \textcolor{green}{the}] unemployed[\textcolor{red}{$\emptyset$} $\rightarrow$ \textcolor{green}{:}] "이태백"[\textcolor{red}{$\emptyset$} $\rightarrow$ \textcolor{green}{,}] "백수"[\textcolor{red}{$\emptyset$} $\rightarrow$ \textcolor{green}{,}] "백조"\end{CJK}

        \emph{\textbf{Proposed Correction 1:}}\begin{CJK}{UTF8}{mj} We have many [\textcolor{red}{New} $\rightarrow$ \textcolor{green}{new}] words for [\textcolor{red}{$\emptyset$} $\rightarrow$ \textcolor{green}{the}] unemployed[\textcolor{red}{$\emptyset$} $\rightarrow$ \textcolor{green}{:}] "이태백"[\textcolor{red}{$\emptyset$} $\rightarrow$ \textcolor{green}{,}] "백수"[\textcolor{red}{$\emptyset$} $\rightarrow$ \textcolor{green}{,}] "백조"\end{CJK}
        \vspace{15pt}
        \\
        \emph{\textbf{Gold Correction 2:}}\begin{CJK}{UTF8}{min} [\textcolor{red}{I and my girlfriend} $\rightarrow$ \textcolor{green}{My girlfriend and I}] looked [\textcolor{red}{$\emptyset$} $\rightarrow$ \textcolor{green}{at a}] picture called "無原罪の聖母" (Immaculate Conception).\end{CJK}

        \emph{\textbf{Proposed Correction 2:}}\begin{CJK}{UTF8}{min} [\textcolor{red}{I and my girlfriend} $\rightarrow$ \textcolor{green}{My girlfriend and}] looked [\textcolor{red}{$\emptyset$} $\rightarrow$ \textcolor{green}{at a}] picture called "無原罪の聖母" (Immaculate Conception).\end{CJK}
        \vspace{10pt}
        \\
        \emph{\textbf{Gold Correction 3}}: \begin{CJK}{UTF8}{min}If he [\textcolor{red}{was a} $\rightarrow$ \textcolor{green}{were}] Japanese, I suppose I [\textcolor{red}{replied} $\rightarrow$ \textcolor{green}{would reply}] like this: "ああ 、駅ならこの道を真っすぐ行けばすぐですよ 、800mくらい先です".\end{CJK}
        
        \emph{\textbf{Proposed Correction 3}}: \begin{CJK}{UTF8}{min}If he was [\textcolor{red}{a} $\rightarrow$ \textcolor{green}{$\emptyset$}] Japanese, I suppose I [\textcolor{red}{replied} $\rightarrow$ \textcolor{green}{would reply}] like this: "ああ 、駅ならこの道を真っすぐ行けばすぐですよ 、800mくらい先です".\end{CJK}
        \end{minipage}
    \caption{Three examples of model's proposed corrections from the CSW test set.}
    \label{fig:csw_examples}
\end{figure*}

\subsection{Inference Tweaking and Error Thresholds}
The inference tweaking phase was crucial in tuning the balance between precision and recall. The changes made here, particularly lowering the minimum error thresholds before the model makes an edit, indicated a clear attempt to force the model to make more corrections.\footnote{Implementation details are presented in Appendix \ref{sec:inf_params}.} While this slightly lowered precision on monolingual errors, it significantly enhanced the performance on CSW text.

To determine that the improved performance of our proposed model was not entirely due to the different inference configuration, we conducted a similar grid search for the existing GECToR model. However, instead of using the Stage 3 validation dataset, as we did with our model, we used the CSW test set directly. The highest $F_{0.5}$ achieved by the baseline model was 56.46, providing evidence that our proposed model beats all inference configurations of the previous GECToR system when applied to CSW texts.

\subsection{Synthetic Data Impact}
The synthetic CSW text and error injection methods were central to this project. The resemblance of our synthetic text to real ESL learner data, as shown by the similarity metrics in Table \ref{table:csw-genuine-metrics}, is a testament to the effectiveness of our chosen generation method. The improvements in $F_{0.5}$ scores provide further evidence of this.

Our extended PIE-synthetic dataset aimed to introduce four error types common in ESL students: noun, pronoun, punctuation and word errors. When compared to the monolingual GECToR, our model is stronger in all of these areas.
\footnote{Error type analysis of our model is given in Appendix \ref{sec:model_errant}.} This provides strong evidence that the targeted approach to error injection was successful in boosting the model's ability in these areas.

\section{Conclusion}
The primary aim of this paper was to build a GEC system capable of effectively correcting English errors in CSW text, whilst maintaining competitive performance on monolingual data. To address the scarcity of CSW data, we explored methods of generating synthetic CSW text. We used several CSW metrics to establish that the LLM prompting-based approach was the most capable of generating text resembling the content in our genuine dataset. From there, we used two error injection methods to create the first substantial datasets labelled for CSW GEC. This significantly expanded the training data available. Importantly, it also opened up opportunities for future research in CSW GEC and CSW NLP more generally. We demonstrated the efficacy of our synthetic data generation techniques by training the first GEC model aimed at correcting errors in CSW texts. Our model showed a clear improvement in performance on CSW data, surpassing the SOTA in this area.

\section{Limitations}
This research, while comprehensive, encounters several limitations that highlight areas for potential improvement. One primary limitation lies in the overrepresentation of Japanese in the genuine CSW dataset. This raises questions about the model's applicability to a broader range of language pairs. This is an unfortunate consequence of using a dataset sourced from Lang-8, a Japanese language learning network. Although our method demonstrated that it could generate texts from a wider range of language pairs, it is possible that all CSW data used shows a bias towards Japanese styles of CSW. Such a bias in our system could inadvertently lead to reduced accessibility and effectiveness for ESL learners who CSW with languages other than Japanese. If this system were to be used as an aid in ESL education, steps should be taken to ensure that it does not contribute to existing inequalities present in language learning platforms. English learning tools should be accessible regardless of the student's native language and future work should focus on developing more inclusive datasets to help mitigate these risks. 

Another possible limitation relates to the style of model chosen. The sequence tagging method was selected due to its lower data requirements, but this decision may have constrained the capabilities of the model. Many of the errors typical of ESL students require complex restructuring of the sentence - a notably difficult task for edit-based GEC systems. Although the data needs are more substantial, it is likely that NMT GEC systems may fare better as they are not constrained by a limited vocabulary of edits.

To assess the likeness of our generated CSW text, we introduced several common CSW metrics. Although useful, these metrics are not very sophisticated, and often struggle to accurately capture the nuances of CSW patterns across different subpopulations. These language patterns can have a substantial impact on the optimal approaches to problems across CSW NLP, and hence, the field would benefit from further research in this area. Ideally, we would have conducted a human study to evaluate the quality of our synthetic data. However, given the constraints of the project, it was not possible, and we acknowledge this as a limitation of our work.

Finally, we reported results for a RoBERTa-base GECToR system. Although we also tested other base models, including BERT, DeBERTa and ELECTRA, we did not look at larger models or ensemble systems. Future extensions could explore this area, building upon the observation that larger models or simple voting ensembles can yield better results than the smaller base models \cite{TARNAVASKYI2022}.

In summary, whilst the current work makes significant contributions to the field of GEC for CSW text, these limitations indicate crucial areas for further research and development.

\bibliography{custom}

\appendix

\section{Example LLM Prompt}
\label{sec:llm_prompt}
Figure \ref{fig:llm_prompt} shows an example LLM prompt used to generate synthetic CSW sentences from genuine examples. As we are using a private subset of the Lang-8 dataset, we are not permitted to share any of the CSW texts.

\label{sec:app_llm_prompt}
\begin{figure*}[h!]
Settings: [no prose]\\
For each of the following code-switched sentences, generate a new sentence that uses the same two languages and a similar style of code-switching. The topic should be different. Ensure you use the correct grammar in the English portion of the sentence. Make sure that each sentence contains 2 languages. Only return the sentences and their number. You must follow all of the instructions.\\

For example, given the source sentence and label: \\
1. This food is called \begin{CJK}{UTF8}{min}
``ラーメン''
\end{CJK}.\\
An acceptable answer would be: \\
1. This animal is called a \begin{CJK}{UTF8}{min}
``犬''
\end{CJK}.\\

Do not include any other information in the generated sentences. The 10 real examples are as follows:\\
    
1. [CSW SENTENCE]\\
2. [CSW SENTENCE]\\
3. [CSW SENTENCE]\\
4. [CSW SENTENCE]\\
5. [CSW SENTENCE]\\
6. [CSW SENTENCE]\\
7. [CSW SENTENCE]\\
8. [CSW SENTENCE]\\
9. [CSW SENTENCE]\\
10.[CSW SENTENCE]\\
\caption{An Example LLM Prompt Used to Generate CSW Text}
\label{fig:llm_prompt}
\end{figure*}

\section{Error Type Analysis of SOTA}
\label{sec:errant_analysis}

Table \ref{table:dataset_errors} shows a breakdown of the performance of a single RoBERTa Large-based GECToR system trained purely on monolingual GEC data when applied to two datasets, our genuine CSW dataset and the BEA-2019 \cite{bea-2019_dataset} test set. These datasets are approximately the same size. The model used represents a current near-SOTA single model sequence tagging-based GEC system measured using $F_{0.5}$ on the BEA-2019 test set. For brevity, we have removed categories with a low number of examples in either dataset or where performance is not significantly different.

\begin{table*}[ht]
\begin{center}
\begin{tabular}{l| r r r r| r r r r |r}
\textbf{Category} & \multicolumn{4}{c}{\textbf{BEA-2019 Test}} & \multicolumn{4}{c}{\textbf{Genuine CSW}}& \textbf{$\text{F}_{0.5}$ Decrease} \\
& \textbf{$\text{F}_{0.5}$} & \textbf{TP} & \textbf{FP} & \textbf{FN} &\textbf{$\text{F}_{0.5}$} & \textbf{TP} & \textbf{FP} & \textbf{FN} \\
\hline
DET & 80.45 & 432 & 80 & 205 & 46.27 & 472 & 351 & 1336 & 34.18\\
NOUN & 47.85 & 29 & 16 & 94 & 4.34 & 21 & 147 & 1725 & 43.51\\
ORTH & 75.96 & 201 & 30 & 198 & 36.45 & 181 & 264 & 522 & 39.51\\
OTHER & 39.51 & 113 & 77 & 557 & 3.55 & 39 & 241 & 4333 & 35.96\\
PREP & 75.44 & 263 & 58 & 196 & 39.14 & 241 & 251 & 870 & 36.30\\
PRON & 66.38 & 62 & 19 & 81 & 20.71 & 21 & 32 & 274 & 45.67\\
PUNCT & 80.93 & 786 & 165 & 266 & 0.35 & 1 & 286 & 284 & 80.58\\
VERB & 52.59 & 61 & 27 & 167 & 18.08 & 49 & 77 & 802 & 34.51\\
VERB:FORM & 81.62 & 151 & 30 & 50 & 37.20 & 61 & 90 & 155 & 44.42\\
VERB:SVA & 88.64 & 128 & 14 & 26 & 57.58 & 114 & 79 & 104 & 31.06\\
VERB:TENSE & 65.55 & 145 & 62 & 133 & 33.80 & 116 & 172 & 448 & 31.75\\
WO & 58.08 & 23 & 5 & 63 & 6.33 & 2 & 11 & 104 & 51.75\\
\hline
\end{tabular}
\end{center}
\caption{$F_{0.5}$ Scores, TP, FP, FN, and Differences in $F_{0.5}$ Scores (BEA - CSW) for Different Categories in the BEA-2019 Test Split and our Genuine CSW Dataset.}
\label{table:dataset_errors}
\end{table*}

\section{Training Data Schedule}
\label{sec:train_data}
In this section, we explicitly detail the data used at each stage of the training process.

\paragraph{Stage 1}
For the initial pre-training stage, we used the distilled dataset proposed by the SOTA \cite{TARNAVASKYI2022}. This dataset was constructed by extracting corrections from the monolingual 1BW corpus \cite{1bw_dataset} using the highest performing GECToR ensemble. Through this dataset, we shuffled our PIE-synthetic CSW dataset. We deemed this dataset to be of lower quality than its Rev-GECToR counterpart. Consequently, it was used earlier in the training process. This provided roughly 1,200,000 examples for the initial training phase of which we split between train and validation sets according to a ratio of 19:1. Our synthetic CSW sentences comprised approximately 5.65\% of this dataset. We aimed to keep this percentage small in this phase of the training process to allow the model to first learn to correct errors in monolingual texts. In later stages, we boosted the contribution of the CSW data.

\paragraph{Stage 2}
For the second stage, we shuffled several GEC datasets. These are NUCLE \cite{nucle_dataset}, FCE \cite{fce_dataset}, W\&I Locness \cite{bea-2019_dataset}, Lang-8 \cite{lang8_dataset} and our 2 newly created CSW datasets. As our genuine CSW dataset is a subset of the private Lang-8 corpus, we checked and removed any duplicates. Table \ref{table:stage_2_datasets} shows the overall contributions of each corpus towards the stage 2 dataset. Similar to the previous stage, the data was split into train and validation sets.

\begin{table}[h!]
\begin{center}
\begin{tabular}{l r r}
\textbf{Dataset} & \textbf{Sentences}\\ [0.5ex]
\hline
Lang-8 & 985,683 \textit{(80.54\%)}\\
W\&I Locness & 68,608 \textit{(5.61\%)}\\
NUCLE & 54,258 \textit{(4.43\%)}\\
FCE &  26,929 \textit{(2.20\%)}\\
\hline
Syn-CSW PIE & 70,181 \textit{(5.73\%)}\\
Syn-CSW Rev-GECToR & 18,160 \textit{(1.48\%)}\\
\hline
Total & 1,223,819\\
\hline
\end{tabular}
\end{center}
\caption{Sentence Count and Contribution of Stage 2 Datasets}
\label{table:stage_2_datasets}
\end{table}

\paragraph{Stage 3}
For the final stage, we combined the high quality W\&I Locness dataset with a sampled subset of the genuine CSW data and a sampled subset of the synthetic CSW texts. Again, the stage 3 dataset is split into train and validation sets. The remaining unused subset of the genuine CSW dataset was retained for testing purposes. Table \ref{table:stage_3_datasets} details the contributions to this stage from each dataset.

\begin{table}[h!]
\begin{center}
\begin{tabular}{l r r}
\textbf{Dataset} & \textbf{Sentences} \\ [0.5ex]
\hline
W\&I Locness & 68,608 \textit{(67.23\%)}\\
\hline
Syn-CSW Rev-GECToR & 18,160 \textit{(17.80\%)}\\
Syn-CSW PIE & 10,000 \textit{(9.80\%)}\\
CSW Genuine & 5,279 \textit{(5.17\%)}\\
\hline
Total & 102,047\\
\hline
\end{tabular}
\end{center}
\caption{Sentence Count and Contribution of Stage 3 Datasets}
\label{table:stage_3_datasets}
\end{table}

\section{Inference Hyperparameters}
\label{sec:inf_params}

After training our model, we used the validation dataset from stage 3 to tune 2 inference parameters. These are:
\begin{itemize}
\item additional\_confidence --- This value is added to the probability of the \textit{\$KEEP} token. If this value is high, recall is likely to decrease and precision increase. The grid search found the best value of this to be 0.
\item min\_error\_probability --- For a change to be made to a sentence, the probability of at least one token in the sentence being an error must be higher than the min\_error\_probability. If this value is high, then precision is likely to be higher and recall lower. The grid search found the best value of this to be 0.4.
\end{itemize}

\section{Error Type Analysis of Proposed Model}
\label{sec:model_errant}
By exploring the ERRANT error classifications of our proposed model when applied to the CSW test dataset, we can further explore the effectiveness of our synthetic data in addressing the problematic areas identified in Appendix \ref{sec:errant_analysis}. A breakdown of the precision, recall and $\text{F}_{0.5}$ score for each of the previously identified categories is shown in Table \ref{table:errant_csw_final}.

\begin{table}[ht]
\centering
\begin{tabular}{l  c  c  c}
\textbf{Category} & \textbf{P} & \textbf{R} & $\text{\textbf{F}}_{0.5}$ \\
\hline
NOUN & 0.2857 & 0.0833 & 0.1923\\
PRON & 0.7647 & 0.4643 & 0.6771 \\
PUNCT & 0.7143 & 0.1139 & 0.3460 \\
WO & 0.7778 & 0.2000 & 0.4930 \\
\hline
\end{tabular}
\caption{Precision (P), Recall (R) and $F_{0.5}$ Score of Our Proposed Model for Targeted Error Types in the Genuine CSW Test Dataset}
\label{table:errant_csw_final}
\end{table}


\end{document}